\documentclass[letterpaper,10pt]{article}
\usepackage[preprint]{corl_2026}
\usepackage{amsmath,amssymb}
\usepackage{graphicx}
\usepackage{booktabs}
\usepackage{multirow}
\usepackage{array}
\usepackage{placeins}
\usepackage{capt-of}
\usepackage{wrapfig}
\usepackage{marvosym}
\usepackage{xcolor}
\newcommand{\tablecaptiongap}{6pt}
\hypersetup{
  hidelinks,
  pdftitle={Efficient-WAM: A 1B-Parameter World-Action Model with Low-Cost Future Imagination},
  pdfauthor={Jiajun Li, Tiecheng Guo, Yifan Ye, Rongyu Zhang, Xiaowei Chi, Qianpu Sun, Ying Li, Yunfan Lou, Jinqiao Wang, Zhihe Lu, Meng Guo, Shanghang Zhang},
  pdfsubject={Preprint}
}

\newif\ifanonymous
\anonymousfalse
\title{Efficient-WAM: A 1B-Parameter World-Action Model with Low-Cost Future Imagination}
\author{
{\fontsize{10.5}{12.5}\selectfont\bfseries Jiajun Li\textsuperscript{1,*} \quad
Tiecheng Guo\textsuperscript{2,*} \quad
Yifan Ye\textsuperscript{2,*} \quad
Rongyu Zhang\textsuperscript{2}} \\[0.2em]
{\fontsize{10.5}{12.5}\selectfont\bfseries Xiaowei Chi\textsuperscript{3,\(\ddagger\)} \quad
Qianpu Sun\textsuperscript{2} \quad
Ying Li\textsuperscript{2} \quad
Yunfan Lou\textsuperscript{2}} \\[0.2em]
{\fontsize{10.5}{12.5}\selectfont\bfseries Jinqiao Wang\textsuperscript{4} \quad
Zhihe Lu\textsuperscript{5} \quad
Meng Guo\textsuperscript{2} \quad
Shanghang Zhang\textsuperscript{2,\Letter}} \\[0.35em]
{\normalsize\normalfont \textsuperscript{1}The University of Hong Kong \quad
\textsuperscript{2}Peking University \quad
\textsuperscript{3}Muka Robotics} \\[0.18em]
{\normalsize\normalfont \textsuperscript{4}Institute of Automation, Chinese Academy of Sciences \quad
\textsuperscript{5}Nanjing University} \\[0.18em]
{\normalsize\normalfont \textsuperscript{*}Equal contribution \quad
\textsuperscript{\(\ddagger\)}Project lead \quad
\textsuperscript{\Letter}Correspondence: shanghang@pku.edu.cn} \\[0.18em]
{\normalsize\normalfont Project page: \href{https://efficientwam.github.io/}{\textcolor{blue}{https://efficientwam.github.io/}}}
}

\begin{document}
\maketitle
\raggedbottom
\enlargethispage{3\baselineskip}

\begin{abstract}
World-Action Models (WAMs) have emerged as a promising paradigm for embodied control by coupling future visual prediction with action generation. However, many existing WAMs rely on photorealistic future prediction, which incurs high inference latency and makes real-time robot deployment difficult. We introduce Efficient-WAM, a World-Action Model motivated by the idea that a compact future branch can still support effective action generation, even at lower visual fidelity. Efficient-WAM improves inference efficiency via a compact video expert transferred from WAN-2.2-5B, token-sparse video latents, and asymmetric video-action denoising that allocates fewer sampling steps to video than to actions. Instead of prioritizing visual fidelity, Efficient-WAM treats future video prediction as a compact guidance signal for action generation. Experiments on RoboTwin 2.0 and real-world manipulation tasks show that Efficient-WAM maintains strong action performance despite visibly coarse future predictions. Our 1B-parameter model achieves 98 ms per action chunk during physical deployment, over 30$\times$ faster than the Motus baseline with comparable task success.
\end{abstract}


\vspace{0.7em}
\begin{minipage}{\linewidth}
  \centering
  \includegraphics[width=\linewidth]{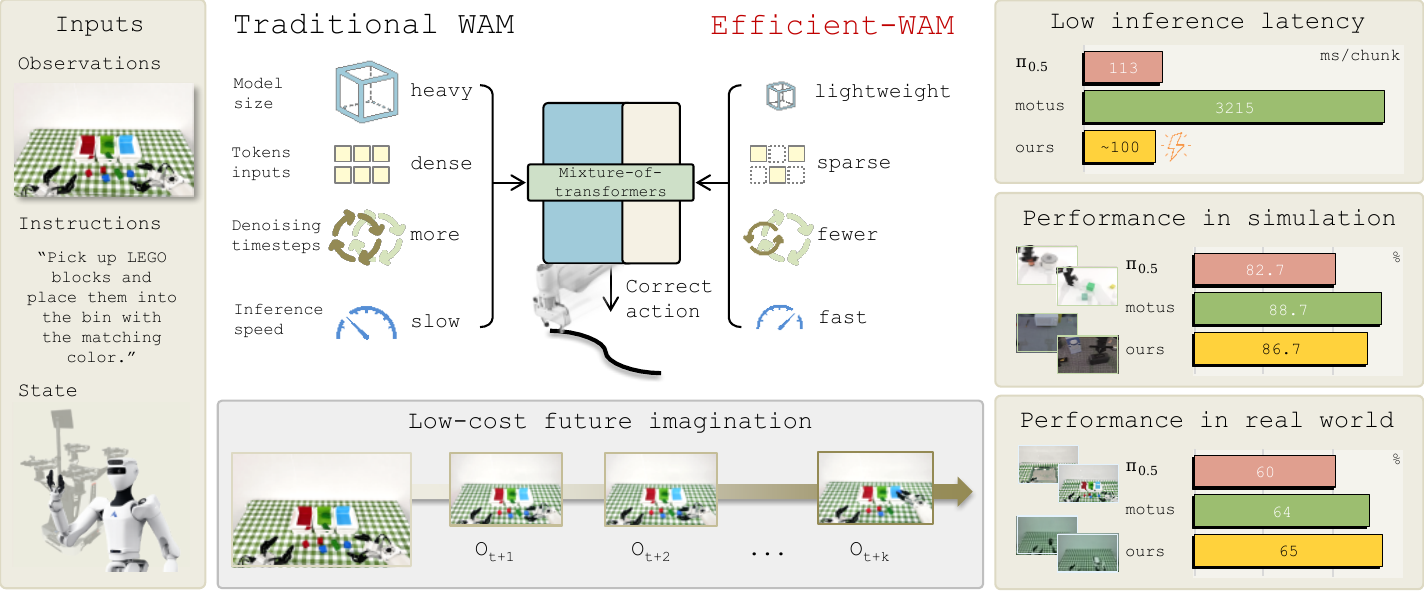}
  \captionof{figure}{\textbf{Overview of Efficient-WAM.} Efficient-WAM reduces the cost of future imagination while maintaining strong task success despite coarse video predictions. It uses a compact video expert, low-resolution future latents, and asymmetric video-action denoising for low-latency deployment.}
  \label{fig:teaser}
\end{minipage}
\newpage

\section{Introduction}
\label{sec:introduction}

Robot control requires understanding how the physical scene will evolve during interaction. World-Action Models (WAMs)~\cite{wang2026worldactionmodels} address this by coupling future video prediction~\cite{finn2017deepvisualforesight, hu2025vpp, wu2024gr1} with action generation~\cite{du2023unipi, li2025uva, kim2026cosmospolicy, ye2026dreamzero}. By predicting how observations change over time, WAMs embed rich physical dynamics and world priors into the control policy, making them a promising robot learning paradigm. Yet many WAMs rely on large video generators~\cite{wu2024gr1, kim2026cosmospolicy}, bringing the computational cost of high-fidelity prediction into the control loop. The resulting latency and hardware demands make real-time deployment challenging.

A different picture is taking shape. Photorealistic future prediction is not always necessary for effective control. For example, VPP~\cite{hu2025vpp} shows that action generation remains effective even when the denoising process is reduced to a single step, while Fast-WAM~\cite{yuan2026fastwam} demonstrates that WAMs can remain competitive even when explicit future generation is skipped during inference. Such tolerance to reduced video computation suggests room to rethink the design of the future predictor itself. We therefore propose Efficient-WAM (Fig.~\ref{fig:teaser}), asking whether a streamlined video branch can retain useful future guidance even at reduced visual fidelity.

Our core idea is to make the video branch smaller and more efficient within a Mixture-of-Transformers (MoT) framework~\cite{bi2025motus, li2026causalworldmodeling} through structured pruning and knowledge transfer from the foundation model WAN-2.2-5B~\cite{wanteam2025wan}. The goal is to bring the benefits of large-scale video pretraining to a smaller model that can run within a robot's inference budget. Once this knowledge has been transferred to a compact backbone, two complementary accelerations follow naturally. First, what the robot observes and what it imagines need not have the same visual fidelity. We keep current observations at high resolution while allowing future predictions to be coarser and less costly. Second, imagining the future need not take as many sampling steps as refining an action. A short video schedule supplies features for continued action refinement, reducing the cost of repeated video prediction. Together, these choices make future imagination less expensive while retaining its role in action generation.

Early-exit and dynamic layer-skipping methods~\cite{yue2024deervla,yang2026dyslvla} mainly reduce policy-backbone computation, while single-step denoising and diffusion-policy distillation methods~\cite{song2023consistencymodels,prasad2024consistencypolicy,wang2025onedp} accelerate action sampling. By contrast, we develop a unified WAM framework that systematically addresses three factors governing video-side inference cost: model size, visual-token count, and denoising steps. The ablations measure their individual effects, and real-world deployment evaluates the combined design. Together, these results establish that a future branch can be compressed across all three dimensions while retaining useful guidance for control.

Efficient-WAM denotes the compact structural baseline, while Efficient-WAM-RT denotes its real-time deployment variant with all three efficiency mechanisms enabled. We evaluate both variants in simulation and Efficient-WAM-RT on real-world tasks. Efficient-WAM achieves 86.7\% clean and 85.7\% randomized success on RoboTwin 2.0. Efficient-WAM-RT achieves 65.0\% success across five real-world tasks, compared with 64.0\% for Motus. It runs at 98 ms per chunk on an RTX 4090, over 30$\times$ faster than Motus in the same deployment setup. Our core contributions are:

\begin{itemize}
  \setlength{\itemsep}{0.2em}
  \setlength{\parsep}{0pt}
  \setlength{\topsep}{0.2em}

  \item We identify costly future prediction as a critical deployment bottleneck of WAMs. Furthermore, our compact future-imagination design shows that effective control can be decoupled from the pursuit of photorealistic video generation.

  \item We decompose video-side inference cost into three factors and accordingly propose Efficient-WAM, combining a compact MoT with structured slicing and distillation, sparse future tokens, and training-free asymmetric denoising. Simulation and real-world experiments validate its efficiency and control performance.
\end{itemize}

\section{Related Works}
\label{sec:related-works}

Recent World-Action Models (WAMs) couple future visual prediction with action generation to inject physical priors into robot policies~\cite{li2025uva,zhu2025uwm,kim2026cosmospolicy,ye2026dreamzero,bi2025motus,ye2026gigaworldpolicy,guo2026xwam}. While these approaches demonstrate the value of future prediction, they often inherit the computationally heavy design of video generators: large backbones, dense visual tokens, and iterative denoising. Recent work revisits how video models support control. Being-H0.7~\cite{luo2026beingh07} avoids raw-pixel prediction, while Fast-WAM~\cite{yuan2026fastwam} skips explicit future generation at inference. AHA-WAM~\cite{cai2026ahawam} reuses planner context across action chunks to reduce repeated planning computation during inference.

Beyond WAMs, efficiency has also been studied in VLA and generative robot policies through compact architectures, quantization, early-exit, token compression, dynamic layer activation, and action-sampling acceleration~\cite{kim2025openvla,yue2024deervla,ye2025teamvla,yang2026dyslvla,guan2025efficientvlasurvey,chi2023diffusionpolicy,song2023consistencymodels,prasad2024consistencypolicy,wang2025onedp}. These methods mainly optimize the policy backbone or action sampler, and are complementary to our focus on the video-imagination bottleneck in WAMs. To preserve world priors after pruning~\cite{fan2020layerdrop}, we draw on knowledge distillation~\cite{hinton2015distilling,jiao2020tinybert}. Our goal is to transfer spatiotemporal knowledge from WAN-2.2-5B~\cite{wanteam2025wan} into a compact video expert for action generation, rather than reproduce its full video-generation capabilities.

\section{Method}
\label{sec:method}

\subsection{Design Formulation}
\label{sec:design-formulation}

A World-Action Model (WAM) explicitly models the joint distribution of future scene evolution and control actions. Given a current observation \(o\), a language instruction \(l\), and a robot state \(s\), the joint prediction objective is formulated as \(p(\mathbf{z}^{v}, a_{1:H} \mid o, l, s)\), where \(\mathbf{z}^{v}\) represents the explicit future visual latents and \(a_{1:H}\) is the action chunk.

We implement this joint prediction with two coupled flow-matching experts. Given noisy future latents \(\mathbf{z}^{v}_{t_v}\) and actions \(a_{1:H}^{t_a}\), the model predicts both velocity fields:

\begin{equation}
\label{eq:joint-future-aware-prediction}
\begin{aligned}
(\hat{\mathbf{u}}^{v},\hat{\mathbf{u}}^{a})
&= F_{\phi,\theta}\!\left(\mathbf{z}^{v}_{t_v},a_{1:H}^{t_a},t_v,t_a;o,l,s\right).
\end{aligned}
\end{equation}

Here, \(\phi\) and \(\theta\) parameterize the video and action experts, and \(t_v,t_a\) are their noise levels. Both outputs depend on both streams through joint attention. Robot state and noisy actions can therefore affect video predictions, while future-video representations inform action generation.

To connect this formulation to efficiency, we focus on the computation required to produce the future representation \(\mathbf{z}^{v}\). Denote the active video model size by \(\mathcal{M}_v\), the future prediction resolution by \(r_v\), the resulting number of future video tokens by \(N_{\text{tok}}^{v}(r_v)\), and the video denoising budget by \(K_v\). For a fixed implementation family, the video-side cost can be described as a function:

\begin{equation}
\label{eq:video-branch-cost}
\mathcal{C}_{\text{video}}
=
\mathcal{F}_{\text{video}}\!\left(
\mathcal{M}_v,\,
N_{\text{tok}}^{v}(r_v),\,
K_v
\right),
\end{equation}

\begin{figure}[t]
    \centering
    \includegraphics[width=\linewidth]{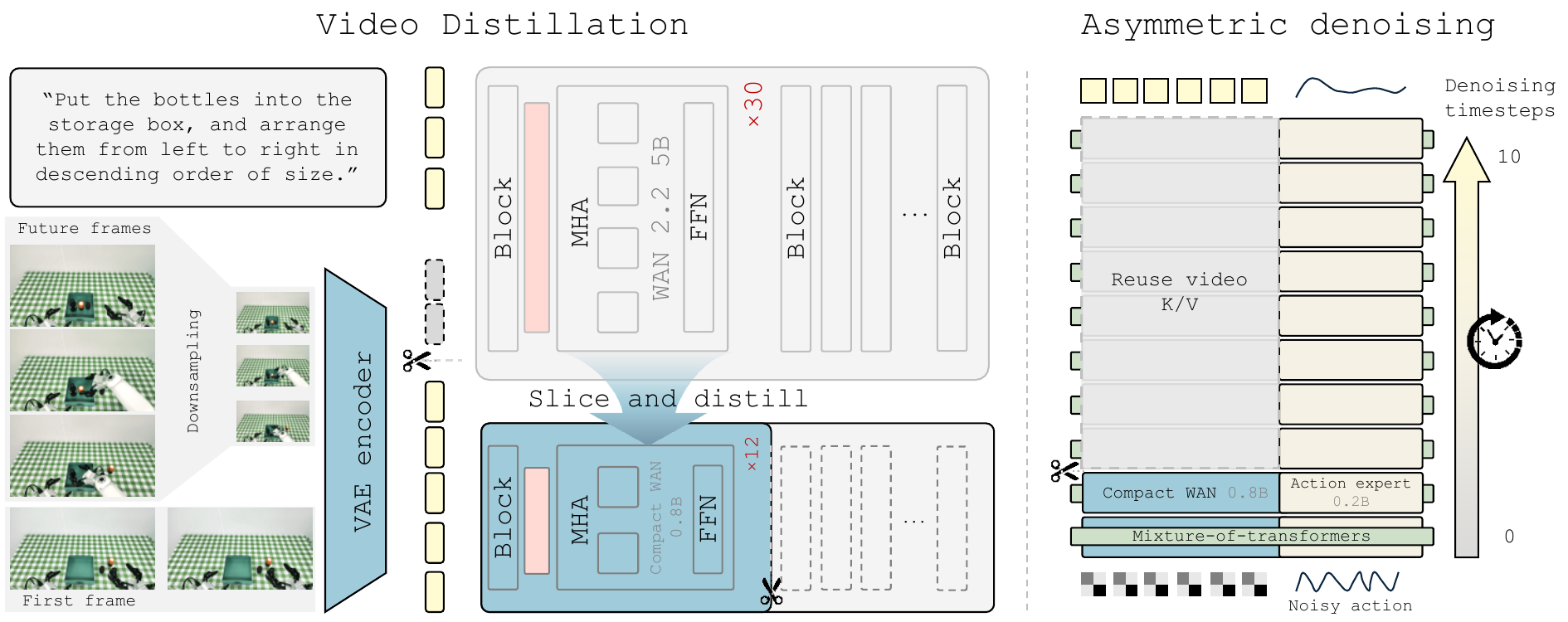}
    \caption{\textbf{Training and inference overview.} Left: Stage~1 transfers a 30-layer video teacher into a 12-layer student through slicing and distillation. Downsampled future frames reduce token count. Right: rows show successive denoising iterations from bottom to top. Both experts run in the first two iterations. The remaining eight update only actions using cached video \(K\) and \(V\) per MoT layer.}
    \label{fig:architecture}
\end{figure}

This abstraction highlights three controllable factors of video-side computation. We reduce them through a compact video expert distilled from WAN-2.2-5B, low-resolution future latents, and asymmetric video-action denoising (Fig.~\ref{fig:architecture}).

\subsection{Compact Architecture with World-Knowledge Transfer}
\label{sec:compact-architecture-with-world-knowledge-transfer}

Large video generators learn rich spatiotemporal priors, but their size makes direct use in low-latency robot control difficult. We transfer these priors from WAN-2.2-5B into a compact video expert through structured initialization and distillation. The aim is to retain the benefits of video pretraining for action generation in a much smaller model. We pair the compact video expert with a dedicated action expert in a Mixture-of-Transformers (MoT) architecture.

To build the 0.8B video expert, we reduce WAN-2.2-5B from 30 to 12 transformer layers and narrow each layer. For depth reduction, we retain the teacher layers used as anchors for two distillation losses. One matches hidden states, and the other matches temporal feature changes. Hidden-state anchors at layers \(1,4,8,14,20,30\) cover shallow-to-deep representations. Motion anchors at \(11,17,23,30\) concentrate temporal-feature supervision in the middle and later layers. The two anchor sets contain nine distinct layers. Finally, we add layers 2, 6, and 26 to fill the gaps and form the 12-layer slice. This gives each distillation anchor a student layer initialized from its teacher counterpart. For width reduction, we select evenly spaced attention heads and FFN channels, preserving each selected head's full dimension. We copy the corresponding rows and columns of teacher weights, with matching slices for biases and normalization parameters. The student has hidden width 2048, FFN width 8192, and 16 attention heads. The two losses are defined in Section~\ref{sec:training-objectives}.

The MoT pairs 12 video layers with 12 action layers of hidden width 768. The instruction enters the video stream through cross-attention. Robot state is embedded as the first token in the action stream, followed by noisy actions. During joint video-action updates, each paired layer projects both streams to matching \(Q\), \(K\), and \(V\) dimensions, concatenates them along the token dimension, and applies bidirectional, unmasked attention. Current-observation, future-video, and state/action tokens can all attend to each other. The attention outputs return to their respective streams, which retain separate projection and FFN parameters.

\subsection{Coarse Future Prediction with Multiscale Latents}
\label{sec:coarse-future-prediction-with-low-resolution-video-tokens}

Predicting future frames at the resolution of the current observation creates a dense video-token sequence. Our framework reduces this cost with a multiscale video-latent layout. Each chunk uses the latest observation, with no visual memory carried across chunks. In all experiments, the current observation is encoded at \(384 \times 320\). Efficient-WAM also predicts future frames at \(384 \times 320\). Efficient-WAM-RT downsamples target future frames to \(192 \times 160\) before VAE encoding. This reduces the number of future tokens to one quarter at a fixed prediction horizon. For Efficient-WAM-RT, the current and future latents are patchified separately and concatenated, with positional encodings based on their respective grids. During inference, the current latent remains a fixed input condition, while future latents are initialized from noise. Joint video-action attention operates on this multiscale sequence. The layout reduces future-token count and attention cost while preserving the input resolution of the current observation.

\subsection{Asymmetric Video-Action Denoising}
\label{sec:asymmetric-video-action-denoising}

Joint video-action denoising runs both experts at every sampling step. The video branch accounts for much of this computation. We allocate separate video and action budgets to reduce video forward passes while retaining more action-refinement steps.

We use training-free asymmetric denoising with independent video and action schedules. For a two-step video budget and a five- or ten-step action budget, we place the two joint updates at the first two action iterations. The video schedule spans the full noise-to-data interval in two steps. Each joint forward caches the video keys and values used at every MoT layer. After the second update, only the action stream runs: its queries attend to cached video keys/values and the current action keys/values. The cache contains intermediate denoiser features, is refreshed at each joint update, and is reset for each new action chunk. No video forward passes are needed during the remaining action steps.

\subsection{Training Objectives}
\label{sec:training-objectives}

Efficient-WAM trains both branches via conditional flow matching~\cite{lipman2023flowmatching,liu2023rectifiedflow}. Let \(\mathbf{x}_0\) denote clean future latents or actions, and \(\mathbf{x}_1 \sim \mathcal{N}(0,I)\). We define \(\mathbf{x}_t = (1-t)\mathbf{x}_0 + t\mathbf{x}_1\) with target velocity \(\mathbf{u}_t = \mathbf{x}_1 - \mathbf{x}_0\). Video and action noise levels are sampled independently during joint training. Each branch uses the loss:
\begin{equation}
\label{eq:unified-flow-loss}
\mathcal{L}_{\text{FM}} = \mathbb{E} \left[ \operatorname{MSE}(\hat{\mathbf{u}},\mathbf{u}_t) \right].
\end{equation}
Here, MSE denotes elementwise mean squared error, and \(\hat{\mathbf{u}}\) is the predicted velocity in Eq.~\eqref{eq:joint-future-aware-prediction}. The expectation covers data, noise, and noise levels. Sampling proceeds from \(t=1\) to \(t=0\).

\textbf{Stage~1: Video distillation.} Teacher and student share noisy video inputs, noise levels, and conditioning. We freeze the teacher and train the compact video expert and its projectors. Teacher features undergo LayerNorm and fixed PCA projection to 256 dimensions. Student LayerNorm--linear projectors are learned separately for each anchor and loss. Let \(\tilde h_{S,n}^{\ell}\) and \(\tilde h_{T,n}^{\tau(\ell)}\) denote projected student and teacher features at matched visual token \(n\), where \(\tau\) maps student to teacher layers. For motion distillation, we spatially average motion-projected tokens within each latent frame to obtain \(m_f\), then compute \(\Delta m_f=m_{f+1}-m_f\). With cosine distance \(d(x,y)=1-\cos(x,y)\), the losses are:
\begin{equation}
\label{eq:distillation-losses}
\begin{aligned}
\mathcal{L}_{\text{hid}}
&= \mathbb{E}_{\ell\in\mathcal{A}_h,n}\,d\!\left(\tilde h_{S,n}^{\ell},\tilde h_{T,n}^{\tau(\ell)}\right),\\
\mathcal{L}_{\text{mot}}
&= \mathbb{E}_{\ell\in\mathcal{A}_m,f}\,d\!\left(\Delta m_{S,f}^{\ell},\Delta m_{T,f}^{\tau(\ell)}\right).
\end{aligned}
\end{equation}
The student anchor sets \(\mathcal{A}_h\) and \(\mathcal{A}_m\) correspond to the teacher anchors in Section~\ref{sec:compact-architecture-with-world-knowledge-transfer}. These losses average over anchors and tokens or adjacent frame pairs. In Stage~1, all loss expectations include normalized, noise-dependent batch weighting specific to each loss. The objective is:
\begin{equation}
\label{eq:stage1-objective}
\mathcal{L}_{\text{stage-1}}=\mathcal{L}_{\text{video-FM}}
+\lambda_d(\mathcal{L}_{\text{hid}}+\mathcal{L}_{\text{mot}}).
\end{equation}
The video flow-matching coefficient remains 1, while \(\lambda_d\) decreases from \(0.2\) to \(0.1\) to \(0\). The teacher and distillation projectors are used only in Stage~1.

\textbf{Stage~2: Action training.} We attach the action expert and optimize \(\mathcal{L}_{\text{joint}}=\mathcal{L}_{\text{action-FM}}+0.01\mathcal{L}_{\text{video-FM}}\) with video parameters frozen. Through bidirectional joint attention, video queries attend to action keys/values, allowing the video loss to backpropagate through frozen video operations to the action expert: \(\Delta\phi=0\), but \(\nabla_\theta\mathcal{L}_{\text{video-FM}}\) can be nonzero.

\textbf{Stage~3: Joint refinement.} We unfreeze the video transformer and train it together with the action expert using \(\mathcal{L}_{\text{joint}}\).

All stages use AdamW with weight decay \(10^{-3}\), a global batch size of \(128\) (\(16\times8\)), cosine learning-rate decay, and bf16 precision. Stages~1--2 use a learning rate of \(5\times10^{-5}\). Stage~3 uses \(10^{-5}\) for the video transformer and \(5\times10^{-5}\) for the action expert.

\section{Experiments}
\label{sec:experiments}

\subsection{Experimental Setup and Model Variants}
\label{sec:experimental-setup}

\begingroup
\widowpenalty=0
\textbf{Efficient-WAM} serves as our structural baseline, isolating the compact architecture (Section~\ref{sec:compact-architecture-with-world-knowledge-transfer}). It uses \(384 \times 320\) for both current observations and future frames, with 10 denoising steps for both video and action. The model \mbox{combines} a 0.8B video expert with an action expert, totaling approximately 1B parameters across the two experts.
\par\endgroup

\textbf{Efficient-WAM-RT} retains \(384 \times 320\) current observations, downsamples future frames to \(192 \times 160\), and uses two video steps and ten action steps in both simulation and real-world evaluation.

Both variants predict 16-action chunks for closed-loop control~\cite{zhao2023aloha}. We measure wall-clock latency per policy call identically in simulation and real-world evaluation. We exclude the first warm-up call. Timing starts after image and state preprocessing and ends after action denormalization. It includes current-observation VAE encoding and all video-action denoising. It excludes one-time T5 encoding, whose embedding is cached, as well as predicted-video VAE decoding and saving, CPU transfer, observation acquisition, and robot execution. We use one A800 GPU for simulation and ablations, and one RTX 4090 for real-world evaluation.

\subsection{Evaluation in Simulation Environment}
\label{sec:simulation-results-and-efficiency-analysis}
\begin{wraptable}{r}{0.52\textwidth}
\setlength{\belowcaptionskip}{\tablecaptiongap}
\vspace{-1.3em}

\centering
\footnotesize
\renewcommand{\arraystretch}{0.9}
\caption{\textbf{RoboTwin 2.0 results.} Baseline success rates are from official papers or technical reports.}
\label{tab:robotwin2_results}

\setlength\tabcolsep{4pt}
        \resizebox{\linewidth}{!}{
  \begin{tabular}{lccc}
    \toprule
    \textbf{Method}
    & \textbf{Clean (\%)}
    & \textbf{Random (\%)}
    & \textbf{Params} \\
    \midrule

    \multicolumn{4}{c}{\textit{VLA-based methods}} \\
    \midrule
    $\pi_0$              & 65.9 & 58.4 & 3.3B  \\
    StarVLA-$\alpha$     & 76.8 & 79.1 & 2B    \\
    $\pi_{0.5}$          & 82.7 & 76.8 & 3.3B  \\
    ABot-M0              & 86.1 & 85.1 & 4.2B \\
    LingBot-VLA          & 86.5 & 85.3 & 4B    \\

    \midrule
    \multicolumn{4}{c}{\textit{WAM-based methods}} \\
    \midrule
    UWM                  & 81.7 & 78.6 & 5B    \\
    GigaWorld-Policy     & 86.4 & 85.0 & 5B    \\
    Motus                & 88.7 & 87.0 & 8B    \\
    Fast-WAM             & 91.9 & 91.8 & 6B    \\
    \textbf{Efficient-WAM}    & 86.7 & 85.7 & \multirow{2}{*}{1B} \\
    \textbf{Efficient-WAM-RT} & 83.1 & 82.0 &  \\

    \bottomrule
  \end{tabular}
        }

\vspace{-0.5\intextsep}
\end{wraptable}
We evaluate on RoboTwin 2.0~\cite{chen2025robotwin2}, comprising 50 bimanual manipulation tasks under clean and randomized visual settings. Both Efficient-WAM variants are trained on 2,500 clean and 25,000 randomized demonstrations. We evaluate each variant with 100 trials per task per setting. VLA baselines include \(\pi_0\)~\cite{black2025pi0}, StarVLA-\(\alpha\)~\cite{ye2026starvlaalpha}, \(\pi_{0.5}\)~\cite{black2025pi05}, ABot-M0~\cite{yang2026abotm0}, and LingBot-VLA~\cite{wu2026lingbotvla}. WAM baselines include UWM~\cite{zhu2025uwm}, GigaWorld-Policy~\cite{ye2026gigaworldpolicy}, Motus~\cite{bi2025motus}, and Fast-WAM~\cite{yuan2026fastwam}.

\begin{figure}[htbp]
  \centering
  \includegraphics[width=\linewidth]{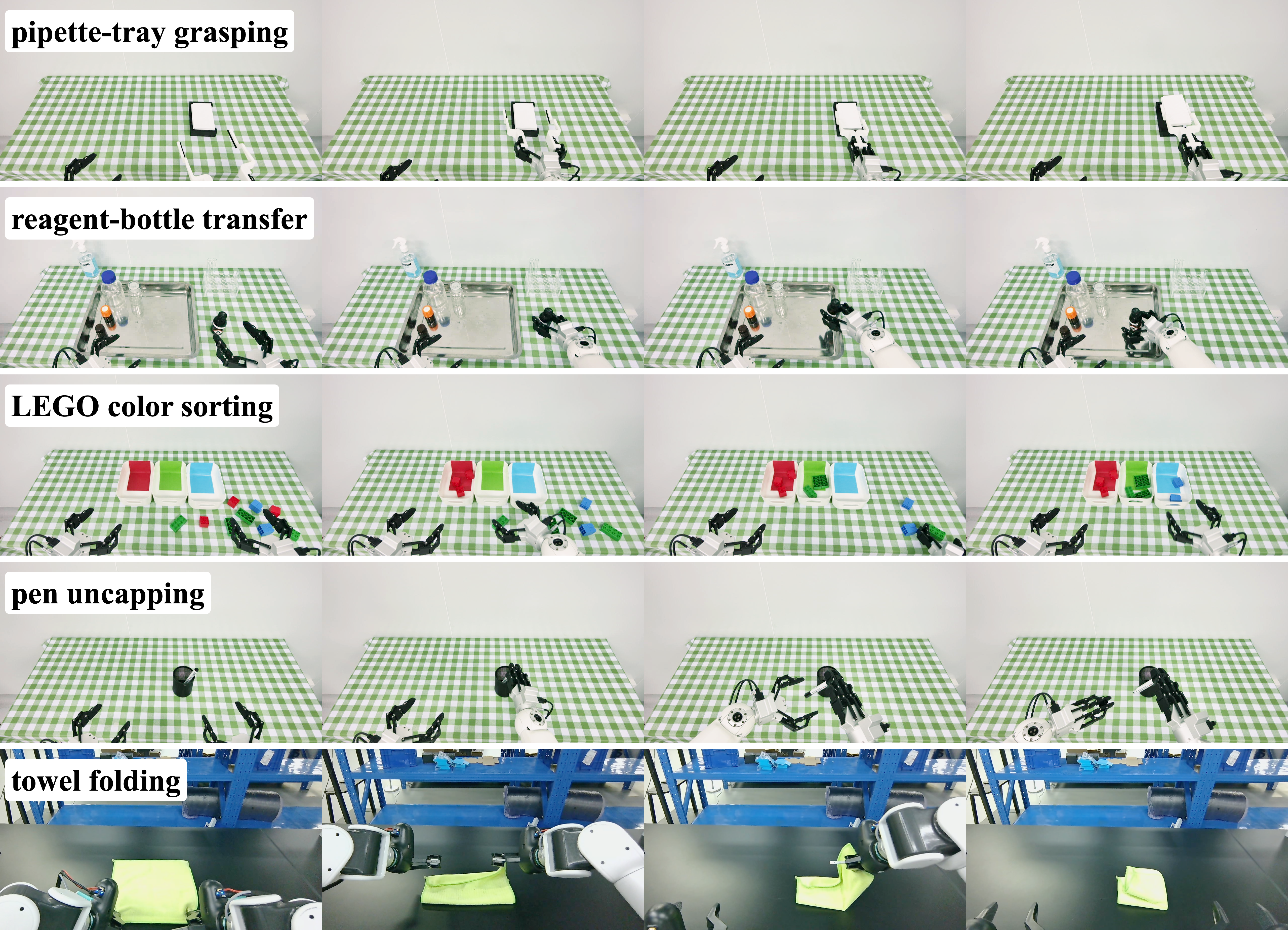}
  \caption{\textbf{Real-world manipulation tasks.} We evaluate five tasks across two dual-arm platforms, covering rigid-object manipulation and deformable-object folding. The towel-folding main-camera view is recorded during autonomous execution for visualization only and is not a policy input.}
  \label{fig:real-world-tasks}
\end{figure}

At 1B parameters, Efficient-WAM achieves 86.7\% clean success and 85.7\% randomized success on RoboTwin 2.0 (Table~\ref{tab:robotwin2_results}), close to the 4B LingBot-VLA and 5B GigaWorld-Policy while trailing the 8B Motus by only 2.0 percentage points in the clean setting. These results show that strong task performance can be retained after model compression. Efficient-WAM-RT achieves lower latency with only a small drop in success, reaching 83.1\% clean and 82.0\% randomized success. It remains above several larger baselines, including $\pi_0$ and StarVLA-$\alpha$. Fast-WAM reports higher success rates, but our model uses about one-sixth the parameters and runs faster with asymmetric denoising (Section~\ref{sec:ablation-studies}). These results highlight the balance between model compactness and task success. Our goal is to retain competitive control performance within a smaller computational budget, making future prediction more practical for robot deployment.

\subsection{Real-World Experiments}
\label{sec:real-world-experiments}

We evaluate Efficient-WAM-RT (Ours), \(\pi_{0.5}\)~\cite{black2025pi05}, and Motus~\cite{bi2025motus} on five real-world tasks across two platforms (Fig.~\ref{fig:real-world-tasks}). The first four use Astribot S1 with RGB views from the head and both wrists and 31-dimensional joint states. Towel folding uses a separate dual-arm humanoid platform with Pika UMI end effectors. For this task, the policy observes only the two end-effector fisheye views and uses UMI's relative-pose state and action representation~\cite{chi2024universal}. We initialize \(\pi_{0.5}\) from its official base checkpoint and Motus from WAN-2.2-5B. Both baselines undergo full-parameter supervised fine-tuning. Each method uses a separate policy per task, trained on the same data: 100 demonstrations per Astribot task and 1,000 UMI demonstrations for towel folding. We run 20 trials per task, with objects reset to randomized, feasible poses and a three-minute limit per trial.

\enlargethispage{\baselineskip}
\textbf{Success criteria.} For all models, we count a trial as successful only if the task requirements below are met within the time limit. Pipette-tray grasping requires lifting the tray from its rack and keeping it level on the end-effector without dropping it or touching its top surface. Reagent-bottle transfer requires lifting and setting down the bottle without drops, abrupt impacts, or collisions with adjacent containers, reflecting laboratory safety concerns~\cite{sun2026labshield}. LEGO color sorting requires placing all 3--5 blocks in each trial into color-matched containers, leaving none on the table. Pen uncapping requires one arm to stabilize the pen body while the other removes the cap, with the pen holder kept upright and neither the pen nor the cap dropped. Towel folding requires two successive folds, first forward with both arms, then leftward with the right arm. All grasps must be successful, with each fold reducing the towel to roughly half its previous size.

\begin{wraptable}{r}{0.52\textwidth}
\setlength{\belowcaptionskip}{\tablecaptiongap}
\vspace{-0.6em}

\centering
\footnotesize
\renewcommand{\arraystretch}{0.9}
  \caption{\textbf{Real-world evaluation.} The \(\dagger\)-marked task uses a separate robot platform and experimental setup. RTX 4090 latency is reported per chunk and amortized over 16 actions.}
  \label{tab:real-world-results}

\setlength\tabcolsep{4pt}
        \resizebox{\linewidth}{!}{
  \begin{tabular}{lccc}
    \toprule
    Real-World Task & \(\pi_{0.5}\) & Motus & Ours \\
    \midrule
    \textit{pipette-tray grasping} & \textbf{100.0} & 85.0 & 95.0 \\
    \textit{reagent-bottle transfer} & 75.0 & \textbf{80.0} & 75.0 \\
    \textit{LEGO color sorting} & 30.0 & \textbf{65.0} & \textbf{65.0} \\
    \textit{pen uncapping} & 10.0 & 25.0 & \textbf{30.0} \\
    \textit{towel folding}\(^{\dagger}\) & \textbf{85.0} & 65.0 & 60.0 \\
    \midrule
    Avg. Success Rate(\%) & 60.0 & 64.0 & \textbf{65.0} \\
    Avg. Lat. per Chunk(ms) & 113.0 & 3215.0 & \textbf{98.0} \\
    Amortized Lat. per Action(ms) & 7.1 & 200.9 & \textbf{6.1} \\
    \bottomrule
  \end{tabular}
        }

\vspace{-0.8\intextsep}
\end{wraptable}

As shown in Table~\ref{tab:real-world-results}, Efficient-WAM-RT achieves a success rate of \(65.0\%\) (65/100 trials) across five real-world tasks and runs at 98 ms per chunk on an RTX 4090. The reported per-action latency of 6.1 ms is calculated as \(98/16\), representing the amortized prediction cost over all 16 actions in each predicted chunk. These results demonstrate low-latency control across the evaluated tasks.

\textbf{Failure analysis.} All three models exhibit grasp misalignment, overlooked objects, and unintended collisions. For \(\pi_{0.5}\), failures include leaving LEGO blocks unsorted and toppling the pen holder. These failures may reflect incomplete task adaptation with 100 demonstrations, despite large-scale pretraining. Its stronger towel-folding performance highlights variation across tasks and setups. For the two WAMs, coupled video-action prediction may make policies sensitive to visual differences between UMI collection and deployment. All towel demonstrations come from a single Pika UMI setup. Despite matching camera models and settings, the robot-mounted cameras produce different images and VAE latents at approximately matched viewpoints. This appearance shift may contribute to unstable behavior during execution of the towel-folding task.

\subsection{Ablation Studies}
\label{sec:ablation-studies}

We first assess the role of future interaction, then isolate the effects of the three video-side efficiency designs through controlled ablations on RoboTwin 2.0.

\paragraph{(a) Effect of Future Prediction}
\begin{wraptable}{r}{0.52\textwidth}
  \setlength{\belowcaptionskip}{\tablecaptiongap}
  \vspace{-1.8em}
  \centering
  \footnotesize
  \renewcommand{\arraystretch}{1.05}
  \setlength{\tabcolsep}{4pt}
  \caption{\textbf{Future-interaction ablation on RoboTwin 2.0.} Entries are clean/randomized success rates (\%). Parentheses give the shared training batch size.}
  \label{tab:fastmask}
  \resizebox{\linewidth}{!}{%
    \begin{tabular}{@{}lcc@{}}
      \toprule
      \textbf{Model (batch size)} & \textbf{Future interaction} & \textbf{FastMask} \\
      \midrule
      WAM-5B (\(16\times16\)) & 86.4 / 85.5 & 76.9 / 73.8 \\
      WAM-5B (\(16\times64\)) & 88.0 / 89.4 & 83.1 / 83.3 \\
      Efficient-WAM (\(16\times8\)) & 86.7 / 85.7 & 65.8 / 65.6 \\
      \bottomrule
    \end{tabular}%
  }
\end{wraptable}
We test the role of future interaction using FastMask, which adapts Fast-WAM's attention mask and inference scheme to our MoT design~\cite{yuan2026fastwam}. It retains video co-training, prevents action tokens from attending to future tokens, and skips future generation at inference. We construct pairs with either the full WAN-2.2-5B or compact Efficient-WAM backbone.\vadjust{\vskip 3pt} Within each pair, backbone and action-expert architectures, layer-wise MoT structure, training data, optimization settings, and evaluation protocol are identical. WAM-5B pairs the full WAN-2.2-5B video expert with a depth/width-matched action expert, using the same MoT interface as Efficient-WAM. All models use \(384\times320\) current observations and ten action-denoising steps. The future-interaction variants additionally predict \(384\times320\) future frames with ten video-denoising steps. All evaluations follow the Section~\ref{sec:simulation-results-and-efficiency-analysis} protocol: 100 trials per task per setting across all 50 tasks.

\begin{table}[!t]
  \centering
  \setlength{\belowcaptionskip}{\tablecaptiongap}
  \caption{\textbf{Ablations on video expert design:} (a) initialization and distillation and (b) future resolution. Success rates are percentages. Parameters count only the video expert. Latency is in ms per chunk on an A800.}
  \label{tab:ablation-compact-video-expert}
  \label{tab:ablation-future-resolution}
  \scriptsize
  \setlength{\tabcolsep}{3pt}
  \renewcommand{\arraystretch}{1.00}
  \resizebox{0.99\columnwidth}{!}{%
    \begin{tabular}{@{}lcccccc@{\hspace{0.8em}}|@{\hspace{0.8em}}lcccc@{}}
      \toprule
      \multicolumn{7}{c}{\textbf{(a) Compact Video Expert}} &
      \multicolumn{5}{c}{\textbf{(b) Future Resolution}} \\
      \cmidrule(lr){1-7}\cmidrule(lr){8-12}
      \textbf{Variant} & \textbf{Init.} & \textbf{Dist.} & \textbf{Clean} & \textbf{Rand.} & \textbf{Params} & \textbf{Lat.} &
      \textbf{Res.} & \textbf{Clean} & \textbf{Rand.} & \textbf{Tokens} & \textbf{Lat.} \\
      \midrule
      Full WAN & Full & -- & 86.4 & 85.5 & 5B & 2013 &
      High & 87 & 86 & 240 & 430 \\
      Compact-random & Random & No & 69 & 68 & 0.8B & 432 &
      Medium & 85 & 84 & 126 & 396 \\
      Compact-sliced & Sliced & No & 82 & 81 & 0.8B & 426 &
      Low & 83 & 82 & 60 & 377 \\
      \textbf{Efficient-WAM} & Sliced & Yes & 87 & 86 & 0.8B & 430 &
       & & & & \\
      \bottomrule
    \end{tabular}%
  }
\end{table}

Table~\ref{tab:fastmask} shows lower success with FastMask in all three comparisons. For Efficient-WAM, clean/randomized success falls from 86.7/85.7\% to 65.8/65.6\%, a drop of 20.9/20.1 percentage points. The full-backbone model shows the same trend at both batch sizes, with a smaller gap in the larger-batch setting. These results favor retaining future interaction under the tested architectures and training settings. However, Fast-WAM still offers a useful insight: video co-training can support effective control without test-time future generation~\cite{yuan2026fastwam}. This works well within its own architecture and training setup, but the success drops above show that the same inference design is not a drop-in replacement for our MoT configurations.
We also measure the original Fast-WAM model on RoboTwin 2.0 with \(H=16\) and ten action-denoising steps. Using the same A800 and timing protocol (Section~\ref{sec:experimental-setup}), we obtain 381.6 ms per chunk.

For the following efficiency ablations, resolution and denoising changes are applied independently to the compact structural baseline. Table~\ref{tab:ablation-compact-video-expert} uses 20 rollouts per task per setting, except for Full WAN, which reuses the 100-rollout results in Table~\ref{tab:fastmask}. Figure~\ref{fig:ablation-asymmetric-denoising} uses 20 rollouts per task on the clean setting. All evaluations cover 50 tasks.

\paragraph{(b) Compact Video Expert}
WAM-5B, as used in (a), serves as the uncompressed baseline. For consistency, all variants use high-resolution futures and symmetric denoising. Table~\ref{tab:ablation-compact-video-expert}a shows that training the compact video expert from random initialization yields 69/68\% clean/randomized success. Layer slicing raises this to 82/81\%, and teacher-guided distillation further improves it to 87/86\%. The distilled model has success rates close to the full-backbone baseline while reducing latency from 2013 ms to 430 ms. These comparisons support pretrained initialization and distillation for the compact expert under our training setup.

\vspace{-0.5em}
\paragraph{(c) Future Resolution}
We vary future resolution in Table~\ref{tab:ablation-future-resolution}b. Reducing token count from 240 to 60 lowers latency from 430 ms to 377 ms with a modest drop in success. Thus, lower-resolution futures retain much of the task performance at reduced computational cost.

\vspace{0.3em}
\noindent
\begin{minipage}[t]{0.48\linewidth}
  \fontsize{10}{12}\selectfont
  \vspace{0pt}
  \textbf{(d) Asymmetric Video-Action Denoising}\par
  \medskip
  We sweep video and action denoising steps, denoted \([T_v,T_a]\), on the clean setting while retaining high-resolution futures. As shown in Figure~\ref{fig:ablation-asymmetric-denoising}, reducing the video budget from \([10,10]\) to \([2,10]\) decreases latency from about 430 ms to 139 ms (a 3.1$\times$ speedup), while success drops by only 0.8 percentage points. Two video updates retain most of the success achieved with ten updates, supporting a smaller video budget during action refinement. With two video updates, we also vary their placement within ten action iterations. We compare the first two iterations with the starred variant, which updates video at iterations one and eight. These placements yield similar inference latency and task success rates, and we use the first two iterations as our default video-update schedule.
\end{minipage}\hfill
\begin{minipage}[t]{0.48\linewidth}
  \fontsize{10}{12}\selectfont
  \vspace{0pt}
  \centering
  \includegraphics[width=\linewidth]{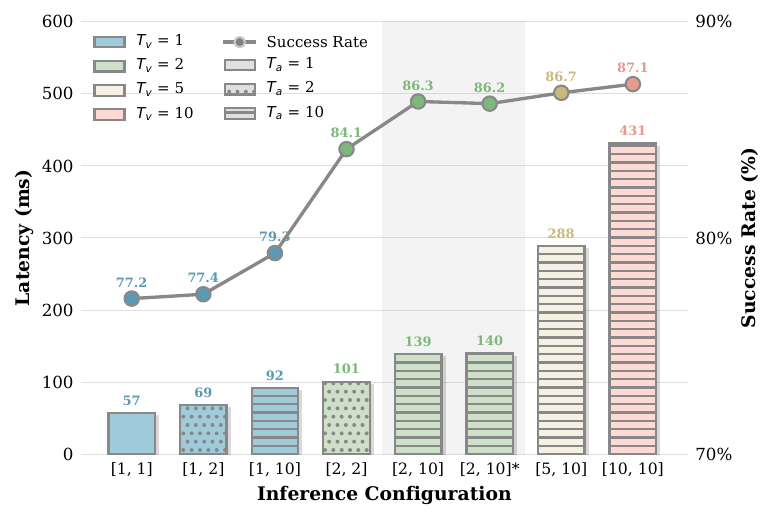}
  \setlength{\abovecaptionskip}{0pt}
  \captionof{figure}{\textbf{Latency--success trade-off on RoboTwin 2.0 (clean).} Shading marks our selected configuration. Updating video in the first two of ten action iterations yields 139.31 ms latency and 86.3\% success. The starred variant updates video at iterations one and eight, yielding 139.97 ms and 86.2\%.}
  \label{fig:ablation-asymmetric-denoising}
\end{minipage}

\FloatBarrier

\newpage
\subsection{Future Prediction Quality}
\label{sec:future-prediction-quality}
\begin{wraptable}{r}{0.52\textwidth}
  \setlength{\belowcaptionskip}{\tablecaptiongap}
  \vspace{-1.8em}
  \centering
  \footnotesize
  \renewcommand{\arraystretch}{1.05}
  \setlength{\tabcolsep}{6pt}
  \caption{\textbf{Future prediction quality.} Video-only evaluation on RoboTwin \textnormal{hanging\_mug}. Higher values indicate better fidelity.}
  \label{tab:future-prediction-quality}
  \begin{tabular}{@{}lcc@{}}
    \toprule
    \textbf{Model} & \textbf{PSNR (dB)} $\uparrow$ & \textbf{SSIM (\%)} $\uparrow$ \\
    \midrule
    WAM-5B & 21.66 & 83.18 \\
    Efficient-WAM & 18.82 & 79.16 \\
    \bottomrule
  \end{tabular}
\end{wraptable}

We compare the video prediction quality of WAM-5B and Efficient-WAM under the same evaluation protocol to assess the effect of compression. We use independent RoboTwin \emph{hanging\_mug} rollouts collected with the official expert policy. We run only the video branches as standalone video generators: the compact 0.8B expert from Efficient-WAM and the full WAN-2.2-5B expert from WAM-5B. No action expert is used during this video-only evaluation, and we retain the original WAM model names for clarity. To quantify the fidelity of the predicted videos, we compute peak signal-to-noise ratio (PSNR) and structural similarity (SSIM).

The lower PSNR and SSIM in Table~\ref{tab:future-prediction-quality} indicate reduced video fidelity for the compact model. Yet its task success remains close to the full-backbone model in the separate control evaluation (Table~\ref{tab:ablation-compact-video-expert}a). Together with the future-interaction comparison in Table~\ref{tab:fastmask}, these results support our insight that compact future prediction can remain useful for action generation despite lower visual fidelity.

\subsection{System-Level Efficiency Analysis}
\label{sec:efficiency-analysis}

The ablations above show how model compression, sparse future tokens, and asymmetric denoising each reduce inference cost. Combining all three, Efficient-WAM-RT achieves 98 ms per action chunk on an RTX 4090, over 30$\times$ faster than Motus at 3215 ms on the same GPU (Table~\ref{tab:real-world-results}). This deployment result is obtained using hardware and experimental settings that differ from those of the A800 ablations, and we therefore report it separately.

\FloatBarrier
\section{Conclusion and Limitations}
\label{sec:conclusion}

We present Efficient-WAM, a 1B-parameter World-Action Model that reduces the cost of future imagination through video-side compression. The ablations show that each design reduces inference cost, while the future-interaction and video-quality evaluations support our central insight: a compact future branch can still benefit action generation despite reduced visual fidelity. Efficient-WAM-RT enables low-latency real-world control with success comparable to Motus. These results jointly demonstrate a practical route to efficient WAM deployment.

\label{sec:limitations}
Nevertheless, low-resolution future prediction trades visual detail for inference speed. Tasks requiring finer spatial precision, such as thread insertion, may still benefit from higher-resolution future representations. This trade-off remains to be evaluated in such settings. The observed visual differences between UMI collection and deployment also motivate further study of robustness to camera variation. Future work will examine these settings in greater detail to develop a more complete understanding of the limits of compact future prediction.

\section*{Acknowledgments}
\ifanonymous\else
This work was supported by the Beijing Natural Science Foundation (L252060).
\fi

\bibliography{reference}
\clearpage
\section*{Appendix}
\appendix

\section{Real-World Evaluation Details}
\label{app:real-world-examples}

This section provides the shared Astribot S1 setup (Fig.~\ref{fig:app-robot-setting}) and representative failure sequences (Fig.~\ref{fig:app-failure-modes}) for the four tasks conducted on that platform. Towel folding uses a separate platform with a different observation and action interface, as described in Section~\ref{sec:real-world-experiments}.

\noindent\begin{minipage}{\linewidth}
\centering
\includegraphics[width=0.58\linewidth]{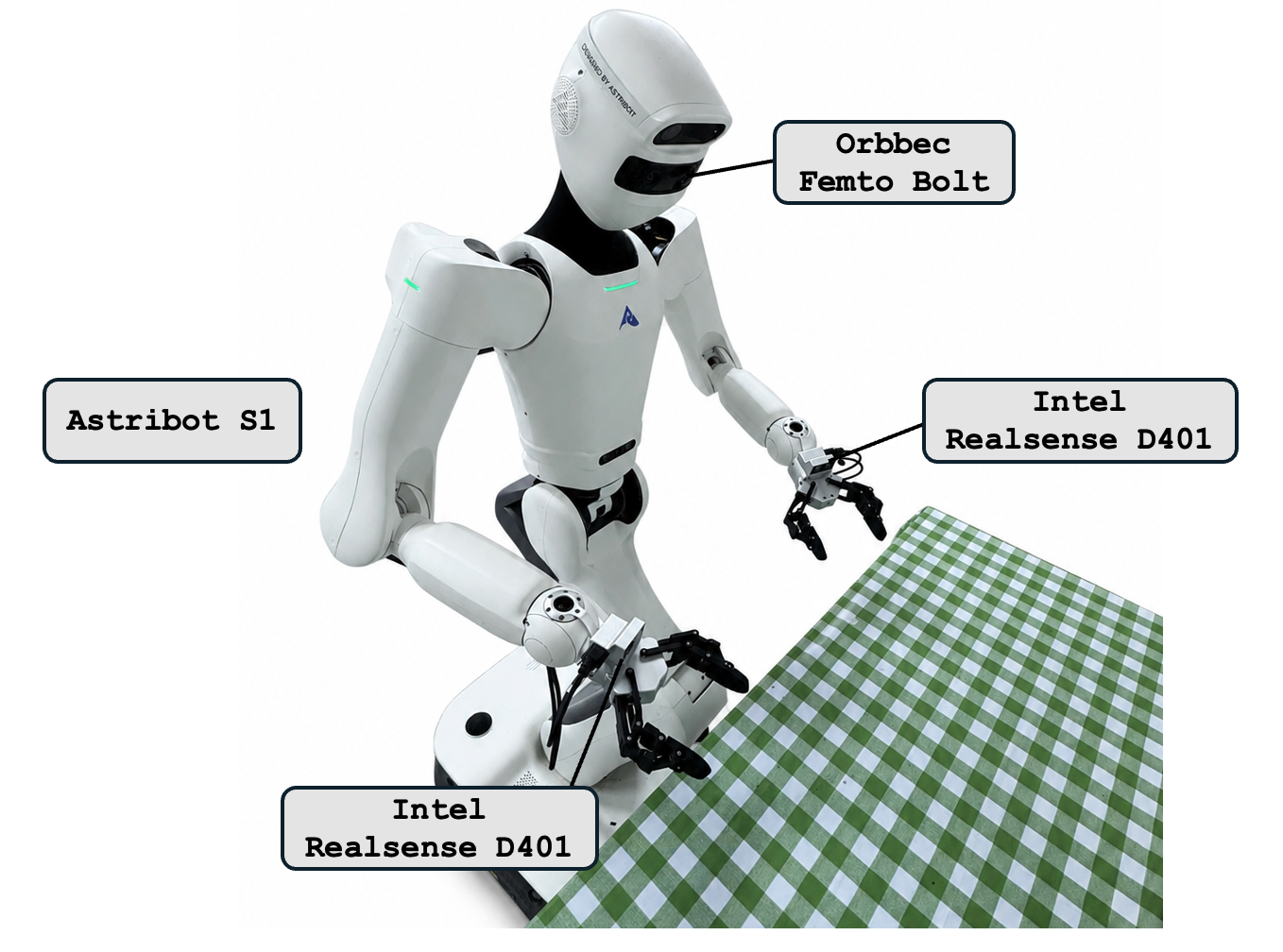}
\captionof{figure}{\textbf{Astribot S1 setup for the first four real-world tasks.} The robot observes the scene through a head camera and cameras mounted near both wrists.}
\label{fig:app-robot-setting}
\end{minipage}

Figure~\ref{fig:app-failure-modes} illustrates the failure categories discussed in the main text. The tray-grasping sequence shows an end-effector misalignment with the rack. In LEGO sorting, one block remains unhandled. The bottle-transfer and pen-uncapping sequences show unintended contact with nearby objects.

\noindent\begin{minipage}{\linewidth}
\centering
\includegraphics[width=\linewidth]{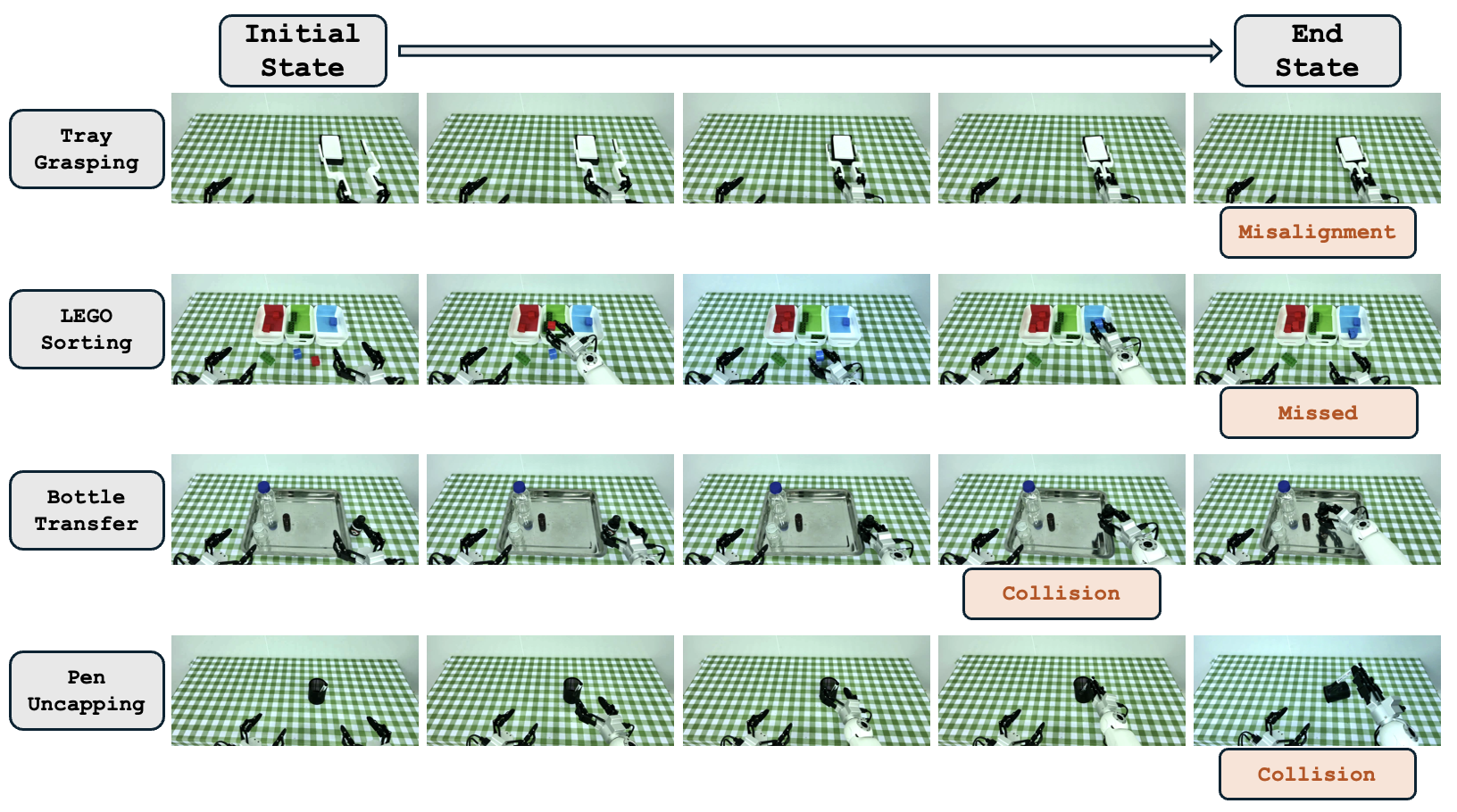}
\captionof{figure}{\textbf{Representative failures in the four Astribot S1 tasks.} Image sequences show tray-grasping misalignment, a missed LEGO block, and collisions during bottle transfer and pen uncapping.}
\label{fig:app-failure-modes}
\end{minipage}

\clearpage

\section{Qualitative Future Prediction Results}
\label{app:qualitative-future}

Figures~\ref{fig:app-fidelity-big} and~\ref{fig:app-fidelity-small} show future predictions from the full-backbone WAM-5B configuration and Efficient-WAM-RT, respectively. The latter combines a compact video expert, lower-resolution future latents, and two video denoising steps. Its predictions are visibly coarser, while the main-text control evaluation shows that this configuration remains effective for action generation (Table~\ref{tab:robotwin2_results}). The examples in this section use Efficient-WAM-RT, while the PSNR and SSIM evaluation in Table~\ref{tab:future-prediction-quality} uses the high-resolution Efficient-WAM video expert under a video-only protocol. Both comparisons use WAM-5B as the reference and show reduced visual fidelity for the compact model.

\vspace{0.6em}
\noindent\begin{minipage}{\linewidth}
\centering
\includegraphics[height=0.35\textheight,keepaspectratio]{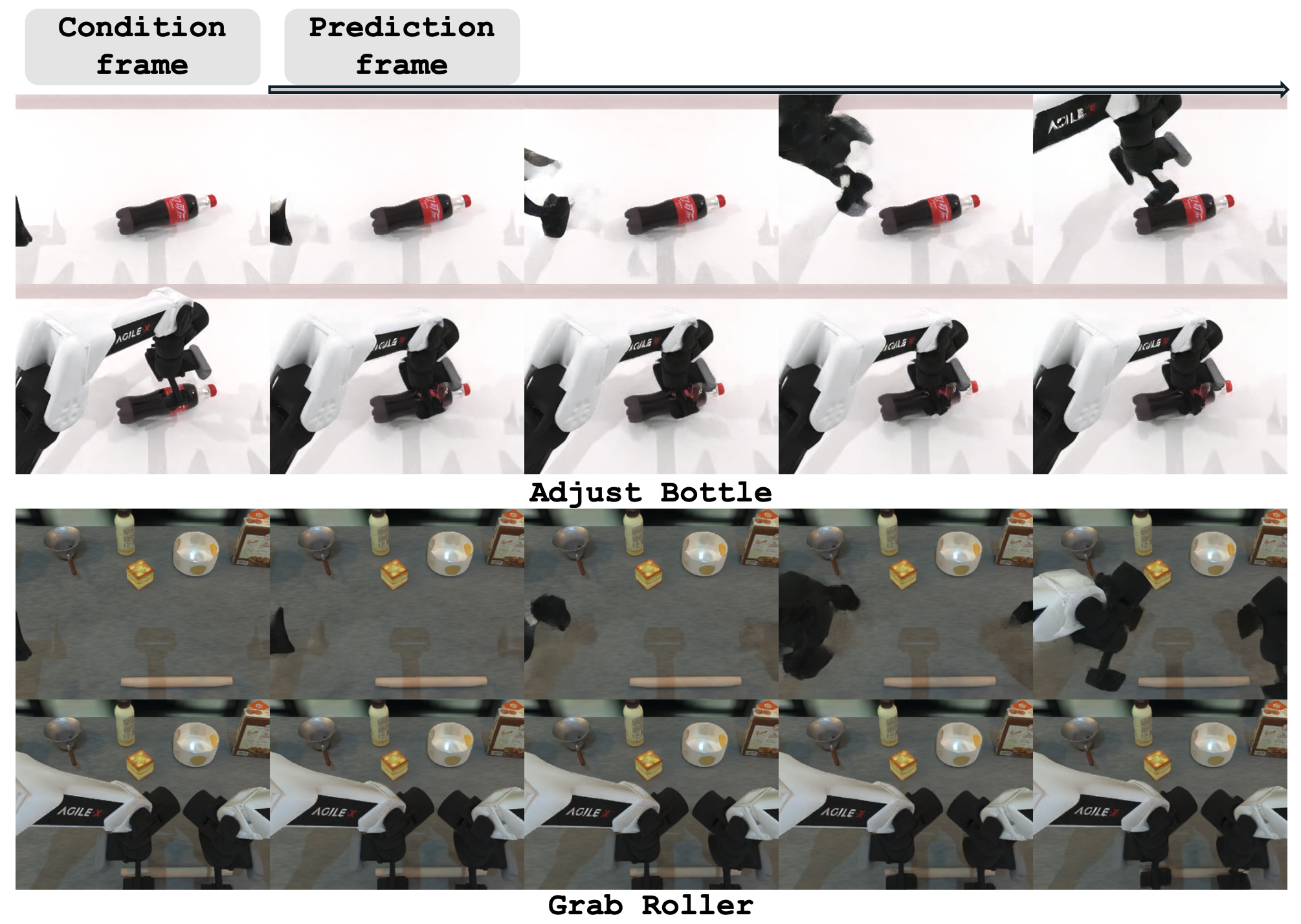}
\captionof{figure}{\textbf{Full-backbone WAM-5B future predictions.} Two RoboTwin tasks.}
\label{fig:app-fidelity-big}
\end{minipage}

\vspace{0.8em}
\noindent\begin{minipage}{\linewidth}
\centering
\includegraphics[height=0.35\textheight,keepaspectratio]{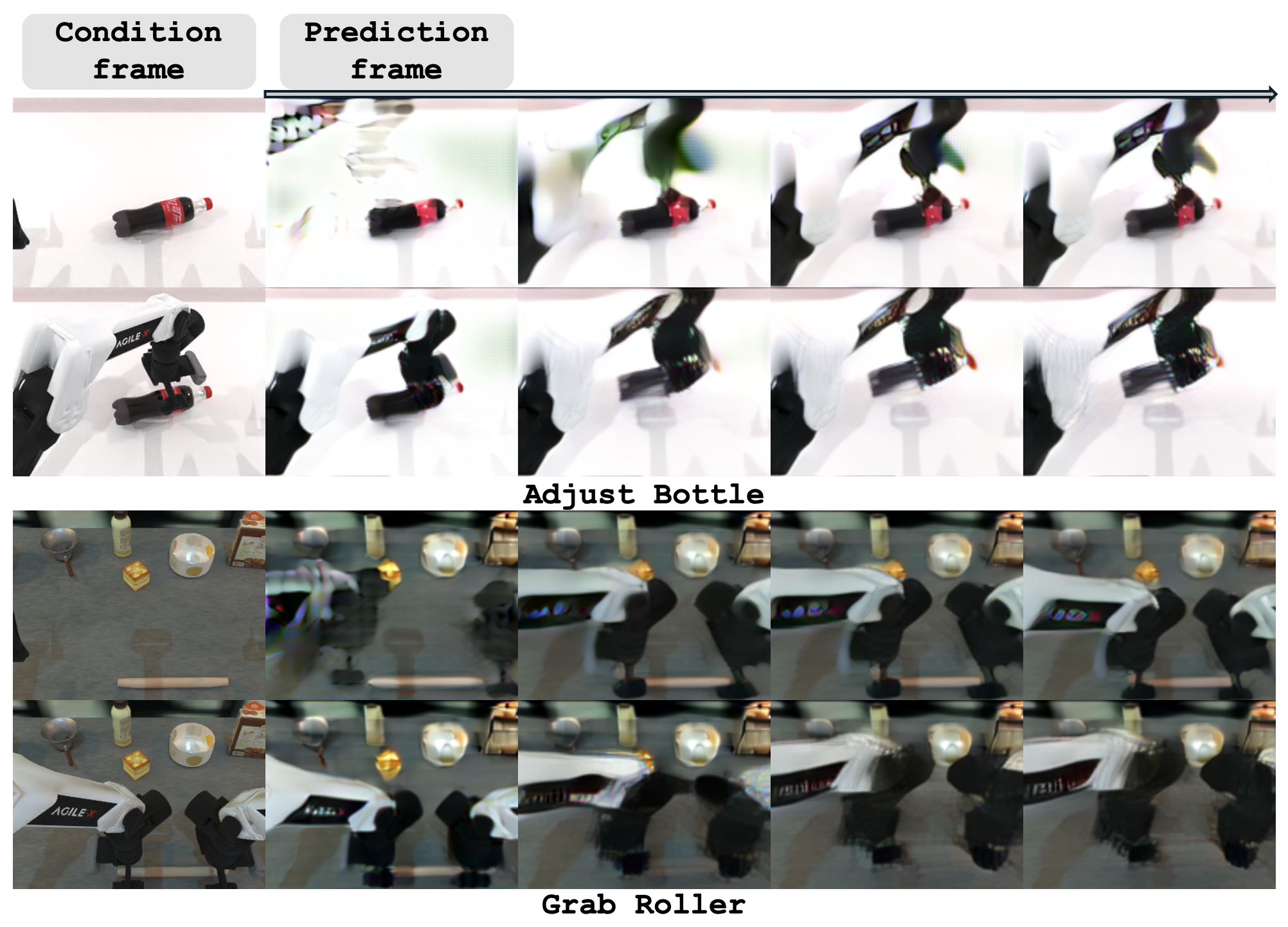}
\captionof{figure}{\textbf{Efficient-WAM-RT future predictions.} Two RoboTwin tasks.}
\label{fig:app-fidelity-small}
\end{minipage}

\clearpage

\section{RoboTwin Detailed Results}
\label{app:robotwin-full-results}

\noindent\begin{minipage}{\linewidth}
\centering
\setlength{\belowcaptionskip}{\tablecaptiongap}
\captionof{table}{\textbf{Per-task RoboTwin 2.0 success rates (\%).} Efficient-WAM and Efficient-WAM-RT are evaluated with 100 trials per task in each setting.}
\label{tab:app-robotwin-full}
\begingroup
\fontsize{10}{10}\selectfont
\setlength{\tabcolsep}{7pt}
\renewcommand{\arraystretch}{1.0}
\begin{tabular}{@{}lcccc@{}}
\toprule
Task & \multicolumn{2}{c}{Efficient-WAM} & \multicolumn{2}{c}{Efficient-WAM-RT} \\
\cmidrule(lr){2-3}\cmidrule(lr){4-5}
 & \textbf{Clean} & \textbf{Rand.} & \textbf{Clean} & \textbf{Rand.} \\
\midrule
adjust bottle & 98 & 98 & 100 & 94 \\
beat block hammer & 94 & 92 & 89 & 82 \\
blocks ranking rgb & 83 & 89 & 82 & 77 \\
blocks ranking size & 50 & 53 & 48 & 45 \\
click alarmclock & 99 & 99 & 100 & 98 \\
click bell & 100 & 100 & 100 & 99 \\
dump bin bigbin & 90 & 90 & 94 & 92 \\
grab roller & 100 & 100 & 100 & 100 \\
handover block & 85 & 78 & 65 & 58 \\
handover mic & 86 & 99 & 82 & 69 \\
hanging mug & 18 & 13 & 23 & 25 \\
lift pot & 96 & 97 & 92 & 90 \\
move can pot & 92 & 94 & 87 & 92 \\
move pillbottle pad & 84 & 89 & 86 & 86 \\
move playingcard away & 96 & 94 & 95 & 86 \\
move stapler pad & 70 & 68 & 75 & 67 \\
open laptop & 90 & 88 & 90 & 96 \\
open microwave & 98 & 98 & 98 & 96 \\
pick diverse bottles & 75 & 67 & 60 & 65 \\
pick dual bottles & 88 & 88 & 67 & 84 \\
place a2b left & 89 & 85 & 90 & 84 \\
place a2b right & 91 & 87 & 91 & 84 \\
place bread basket & 91 & 87 & 87 & 81 \\
place bread skillet & 95 & 92 & 89 & 84 \\
place burger fries & 98 & 100 & 100 & 97 \\
place can basket & 85 & 83 & 88 & 81 \\
place cans plasticbox & 100 & 99 & 99 & 99 \\
place container plate & 99 & 97 & 99 & 99 \\
place dual shoes & 84 & 82 & 79 & 91 \\
place empty cup & 99 & 99 & 94 & 90 \\
place fan & 95 & 89 & 89 & 91 \\
place mouse pad & 84 & 79 & 86 & 76 \\
place object basket & 89 & 87 & 82 & 86 \\
place object scale & 95 & 91 & 92 & 89 \\
place object stand & 93 & 96 & 96 & 92 \\
place phone stand & 82 & 69 & 69 & 67 \\
place shoe & 95 & 97 & 91 & 89 \\
press stapler & 95 & 98 & 99 & 99 \\
put bottles dustbin & 78 & 79 & 32 & 76 \\
put object cabinet & 73 & 60 & 68 & 53 \\
rotate qrcode & 67 & 55 & 70 & 71 \\
scan object & 79 & 79 & 70 & 70 \\
shake bottle & 100 & 99 & 99 & 97 \\
shake bottle horizontally & 100 & 100 & 100 & 97 \\
stack blocks three & 64 & 72 & 65 & 60 \\
stack blocks two & 94 & 98 & 96 & 92 \\
stack bowls three & 67 & 78 & 67 & 78 \\
stack bowls two & 99 & 96 & 87 & 92 \\
stamp seal & 95 & 93 & 95 & 74 \\
turn switch & 69 & 67 & 53 & 60 \\
\midrule
Average & 86.7 & 85.7 & 83.1 & 82.0 \\
\bottomrule
\end{tabular}
\endgroup
\end{minipage}

\clearpage

\noindent\begin{minipage}{\linewidth}
\centering
\setlength{\belowcaptionskip}{\tablecaptiongap}
\captionof{table}{\textbf{Asymmetric-denoising sweep on RoboTwin 2.0 (clean).} Per-task success rates (\%) use 20 rollouts per task. A updates video at iterations 1--2; B at iterations 1 and 8.}
\label{tab:app-asym-denoise-sweep}
\begingroup
\fontsize{8.8}{9.5}\selectfont
\setlength{\tabcolsep}{1.6pt}
\renewcommand{\arraystretch}{1.0}
\begin{tabular}{@{}>{\raggedright\arraybackslash}p{0.235\linewidth}@{\hspace{0.002\linewidth}}*{8}{>{\centering\arraybackslash}p{0.083\linewidth}}@{}}
\toprule
\multirow{2}{*}{\textbf{Task}} & \multicolumn{8}{c}{\textbf{Denoising steps $[T_v,T_a]$}} \\
\cmidrule(lr){2-9}
 & \textbf{$[10,10]$} & \textbf{$[5,10]$} & \textbf{$[2,10]$-A} & \textbf{$[2,10]$-B} & \textbf{$[1,10]$} & \textbf{$[2,2]$} & \textbf{$[1,2]$} & \textbf{$[1,1]$} \\
\midrule
adjust bottle & 95 & 100 & 95 & 100 & 100 & 95 & 100 & 100 \\
beat block hammer & 95 & 100 & 95 & 95 & 85 & 80 & 80 & 80 \\
blocks ranking rgb & 85 & 80 & 85 & 95 & 90 & 95 & 80 & 75 \\
blocks ranking size & 75 & 70 & 35 & 50 & 45 & 50 & 35 & 30 \\
click alarmclock & 100 & 100 & 100 & 100 & 95 & 100 & 100 & 95 \\
click bell & 100 & 100 & 100 & 100 & 100 & 100 & 100 & 100 \\
dump bin bigbin & 90 & 90 & 90 & 90 & 95 & 95 & 100 & 95 \\
grab roller & 100 & 100 & 100 & 100 & 100 & 100 & 100 & 100 \\
handover block & 80 & 95 & 80 & 75 & 65 & 70 & 55 & 35 \\
handover mic & 100 & 95 & 95 & 95 & 60 & 90 & 50 & 50 \\
hanging mug & 20 & 30 & 25 & 5 & 0 & 30 & 25 & 0 \\
lift pot & 100 & 100 & 100 & 100 & 95 & 100 & 95 & 100 \\
move can pot & 95 & 100 & 95 & 90 & 90 & 95 & 85 & 80 \\
move pillbottle pad & 90 & 90 & 85 & 85 & 70 & 85 & 65 & 70 \\
move playingcard away & 95 & 95 & 95 & 90 & 80 & 80 & 75 & 70 \\
move stapler pad & 85 & 75 & 70 & 75 & 80 & 90 & 75 & 80 \\
open laptop & 95 & 95 & 95 & 95 & 95 & 95 & 100 & 100 \\
open microwave & 100 & 100 & 100 & 100 & 100 & 95 & 95 & 80 \\
pick diverse bottles & 70 & 65 & 65 & 70 & 60 & 60 & 60 & 65 \\
pick dual bottles & 85 & 100 & 100 & 95 & 100 & 90 & 85 & 100 \\
place a2b left & 95 & 90 & 95 & 95 & 85 & 95 & 75 & 80 \\
place a2b right & 95 & 95 & 95 & 95 & 85 & 95 & 90 & 95 \\
place bread basket & 85 & 80 & 80 & 80 & 85 & 75 & 80 & 70 \\
place bread skillet & 100 & 95 & 90 & 90 & 80 & 95 & 95 & 95 \\
place burger fries & 85 & 90 & 100 & 90 & 95 & 95 & 95 & 100 \\
place can basket & 95 & 80 & 90 & 65 & 70 & 85 & 85 & 80 \\
place cans plasticbox & 100 & 100 & 100 & 100 & 100 & 90 & 95 & 100 \\
place container plate & 100 & 95 & 95 & 100 & 100 & 95 & 100 & 95 \\
place dual shoes & 70 & 95 & 85 & 75 & 65 & 80 & 65 & 60 \\
place empty cup & 100 & 100 & 100 & 100 & 95 & 100 & 100 & 90 \\
place fan & 90 & 100 & 100 & 100 & 85 & 100 & 80 & 85 \\
place mouse pad & 85 & 75 & 65 & 75 & 65 & 65 & 65 & 65 \\
place object basket & 85 & 85 & 85 & 95 & 80 & 85 & 80 & 80 \\
place object scale & 90 & 95 & 95 & 90 & 90 & 85 & 70 & 85 \\
place object stand & 90 & 90 & 95 & 95 & 95 & 100 & 90 & 80 \\
place phone stand & 80 & 75 & 80 & 70 & 70 & 80 & 45 & 70 \\
place shoe & 85 & 85 & 90 & 90 & 80 & 85 & 75 & 75 \\
press stapler & 95 & 95 & 95 & 95 & 95 & 95 & 95 & 95 \\
put bottles dustbin & 70 & 65 & 70 & 60 & 45 & 55 & 40 & 35 \\
put object cabinet & 85 & 50 & 55 & 75 & 30 & 70 & 25 & 15 \\
rotate qrcode & 90 & 70 & 85 & 90 & 65 & 75 & 85 & 60 \\
scan object & 75 & 70 & 90 & 80 & 55 & 80 & 50 & 70 \\
shake bottle & 100 & 100 & 100 & 100 & 100 & 100 & 100 & 100 \\
shake bottle horizontally & 100 & 100 & 100 & 100 & 95 & 100 & 100 & 100 \\
stack blocks three & 65 & 65 & 65 & 80 & 55 & 50 & 55 & 60 \\
stack blocks two & 95 & 100 & 95 & 95 & 90 & 85 & 75 & 90 \\
stack bowls three & 55 & 70 & 80 & 75 & 50 & 60 & 50 & 75 \\
stack bowls two & 90 & 100 & 95 & 95 & 90 & 90 & 95 & 95 \\
stamp seal & 95 & 95 & 80 & 90 & 80 & 90 & 90 & 85 \\
turn switch & 55 & 50 & 60 & 65 & 85 & 55 & 65 & 70 \\
\midrule
Average & \textbf{87.1} & \underline{86.7} & 86.3 & 86.2 & 79.3 & 84.1 & 77.4 & 77.2 \\
\bottomrule
\end{tabular}
\endgroup
\end{minipage}

\end{document}